\documentclass[11pt,a4paper]{article}
\usepackage[hyperref]{acl2019}
\usepackage{times}
\usepackage{latexsym}

\usepackage{url}

\aclfinalcopy % Uncomment this line for the final submission
\usepackage{graphicx}
\usepackage{amsmath}
\usepackage{bm, upgreek}
\usepackage{amssymb}
\usepackage{array, booktabs}
\usepackage{tabularx}
\usepackage{multirow}
\usepackage{mathtools}
\usepackage{arydshln}
\usepackage{tabularx}
\usepackage{color}
\usepackage{caption}

\newcommand{\Ours}{Dense-Sparse Phrase Index}
\newcommand{\ours}{\textsc{DenSPI}}

\newcommand{\am}{{\bm a}}

\newcommand{\xm}{{\bm x}}
\newcommand{\qm}{{\bm q}}

\newcommand{\ab}{\mathbf{a}}
\newcommand{\bb}{\mathbf{b}}

\newcommand{\db}{\mathbf{d}}

\newcommand{\lb}{\mathbf{l}}
\newcommand{\hb}{\mathbf{h}}

\newcommand{\pb}{\mathbf{p}}
\newcommand{\qb}{\mathbf{q}}

\newcommand{\sbo}{\mathbf{s}}

\newcommand{\xb}{\mathbf{x}}
\newcommand{\Hb}{\mathbf{H}}

\newcommand{\Lb}{\mathbf{L}}

\newcommand{\RR}{\mathbb{R}}

\DeclareMathOperator*{\argmax}{argmax}

\newcommand{\paramp}{\mathrm{Pr}}

\title{Real-Time Open-Domain Question Answering with \\Dense-Sparse Phrase Index}

\author{Minjoon Seo$^{1,5}$\Thanks{ Equal contribution.} \quad Jinhyuk Lee$^{ 6}$\footnotemark[1] \quad Tom Kwiatkowski$^{2}$, \\ {\bf Ankur P. Parikh$^2$ \quad Ali Farhadi$^{1,3,4}$ \quad Hannaneh Hajishirzi$^{1,3}$} \\
  University of Washington$^1$ \quad Google Research$^2$ \quad Allen Institute for AI$^3$ \quad XNOR.AI$^4$ \\
  \quad Clova AI, NAVER$^5$ \quad Korea University$^6$ \\
  {\tt \{minjoon,ali,hannaneh\}@cs.washington.edu} \\
  {\tt \{tomkwiat,aparikh\}@google.com} \quad {\tt jinhyuk\_lee@korea.ac.kr}}

\date{}

\begin{document}
\maketitle
\begin{abstract}
Existing open-domain question answering (QA) models are not suitable for real-time usage because they need to process several long documents on-demand for every input query.
In this paper, we introduce the query-agnostic \emph{indexable} representation of document phrases that can drastically speed up open-domain QA and also allows us to reach long-tail targets. In particular, our dense-sparse phrase encoding effectively captures  syntactic, semantic, and lexical information of the phrases and eliminates the pipeline filtering of context documents.
Leveraging optimization strategies, our model can be trained in a single 4-GPU server and serve entire Wikipedia (up to 60 billion phrases) under 2TB with CPUs only.
Our experiments on SQuAD-Open show that our model is more accurate than DrQA~\cite{drqa} with 6000x reduced computational cost, which translates into at least 58x faster end-to-end inference benchmark on CPUs.\footnote{Visit \url{nlp.cs.washington.edu/denspi} for code \& demo.}
% Existing phrase-level open-domain question answering (QA) systems are not suitable for real-time QA because processing several long documents with a large neural model for every input query is computationally prohibitive. We describe our approach towards super-low-latency open-domain QA purely driven by similarity search.
% To make this possible, we construct an end-to-end hierarchical index where each answer candidate is represented with the concatenation of a document-specific vector and unique phrase-level vector. We show that indexing every phrase in the entire English Wikipedia (3 billion tokens, 5 million documents) is computationally and memory-wise feasible with reasonable computational resources, and that with the help of the recent advancement in deep language contextualization, our system  is on par with or better than previous unconstrained (non-index-based) baselines with more than 36x faster latency on open-domain SQuAD.

\end{abstract}

\section{Introduction}\label{sec:intro}
\begin{figure*}
    \centering
    \includegraphics[width=\linewidth]{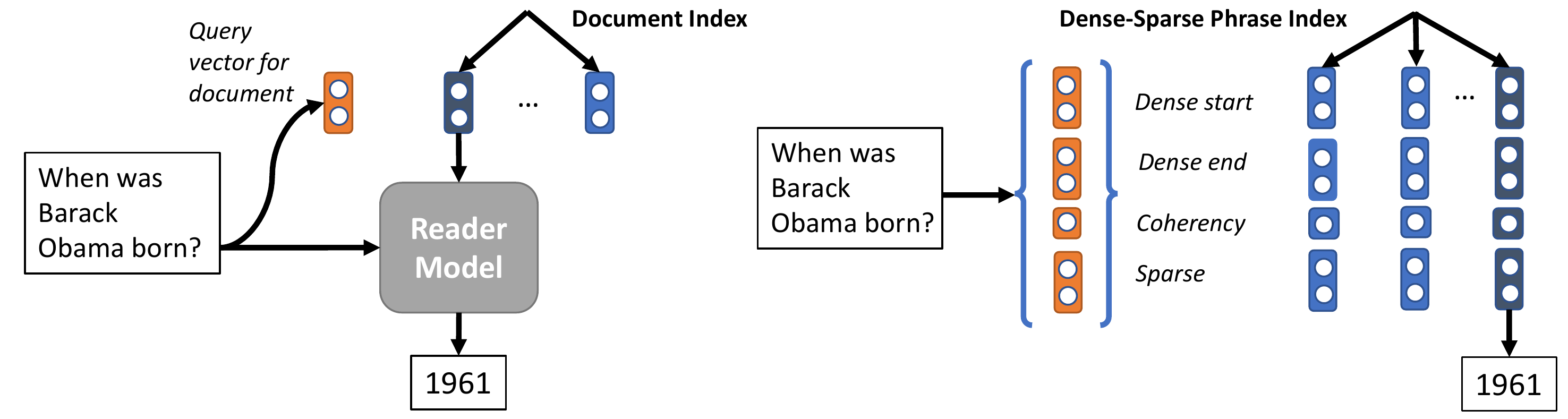}
    \caption{An illustrative comparison between a pipelined QA system, e.g. DrQA~\cite{drqa} (left) and our proposed \Ours\ (right) for open-domain QA, best viewed in color. Dark blue vectors indicate the retrieved items from the index by the query.}
    \label{fig:teaser}
\end{figure*}

Extractive open-domain question answering (QA) is usually referred to the task of answering an arbitrary factoid question (such as ``Where was Barack Obama born?'') from a general web text (such as Wikipedia). 
% Most current popular QA systems~\cite{bidaf,bert} are designed for closed-domain QA where the phrase-level answer to the question can be extracted from a context, which is usually a set of  paragraphs. Despite the great improvement in accuracy, these models are usually not scalable to go beyond multiple documents, which is essential for open-domain QA. 
This is an extension of the reading comprehension task~\cite{squad} of selecting an answer phrase to a question given an evidence document. To make a scalable open-domain QA system,  One can leverage a search engine to filter the web-scale evidence to a few documents, in which the answer span can be extracted using a reading comprehension model~\cite{drqa}. However, the accuracy of the final QA system is bounded by the performance of the search engine due to the pipeline nature of the search process. 
What is more, running a neural reading comprehension model~\cite{bidaf} on a few documents is still computationally costly, since it needs to process the evidence document for every new question at inference time. This often requires multi-GPU-seconds or tens to hundreds of CPU-seconds---BERT~\cite{bert} can process only a few thousand words per second on an Nvidia P40 GPU. %, which is hardly real-time and not suitable for usage on real products.
% \hanna{this should be cleaned up and summarized} 
% Wikipedia has over 3 billion tokens; not only reading the entire corpus is infeasible (millions of GPU-seconds), reducing the search space to 5-10 documents still costs a few seconds on an expensive GPU,  

In this paper, we introduce \Ours\ (\ours), an \emph{indexable} query-agnostic phrase representation model for real-time open-domain QA. The phrase representations are indexed offline using time- and memory-efficient training and storage. During inference time, the input question is mapped to the same representation space, and the phrase with maximum inner product search is retrieved. 
 
Our phrase representation combines both dense and sparse vectors.  Dense vectors are effective for encoding local syntactic and semantic  cues leveraging recent advances in contextualized text encoding~\cite{bert},  while    sparse  vectors  are superior at encoding precise lexical information. 
% that is trained end-to-end for answering factoid questions from Wikipedia and achieves high accuracy compared to the previous pipeline approaches. Our approach extends real-time document retrieval currently employed by search engines to real-time span-level retrieval using end-to-end training of phrase encodings. 
The independent encoding of the document phrases and the question  enables real-time inference; there is no need to re-encode documents for every question. Encoding phrases as a function of their start and end tokens facilitates indexable representations with under 2TB for up to 60 billion phrases in Wikipedia. Moreover, approximate nearest neighbor search on indexable representations allows fast and direct retrieval in a web-scale environment.

Experiments on SQuAD-Open~\cite{drqa} show that \ours\ is comparable or better than most state-of-the-art open-domain QA systems on Wikipedia with up to 6000x reduced computational cost on RAM. In our end-to-end benchmark, this translates into at least 58x faster query inference including disk access time. % that consists of 3 billion tokens and 5 million documents.
%We show that, through end-to-end hierarchical index, indexing billions of phrases in the entire English Wikipedia is computationally and memory-wise feasible, and our model competitive accuracy compared to previous (pipelined) open-domain QA systems with more than 100x faster latency.
%

At the web scale, every detail of the training, indexing, and inference  needs to be carefully designed. %For example, we represent each phrase as a function of its start and end token encodings and store pointers to those vectors instead of storing all possible phrase. 
For reproducibility under an academic setting, we discuss optimization strategies for reducing time and memory usage during each stage in Section~\ref{sec:training}. This enables us to start from scratch and fully deploy the model with a 4-GPU, 64GB memory, 2 TB SATA\footnote{Further speed up is expected when taking a full advantage of a faster interface such as PCIe.} SSD server in a week.

\section{Related Work}\label{sec:related}
\paragraph{Open-domain question answering} Creating a system that can answer an open-domain factoid question has been a significant interest to both academic and industrial communities. The problem is largely approached from two subfields: knowledge base (KB) and text (document) retrieval. Earlier work in large-scale question answering~\cite{kgqa} has focused on answering questions from a structured KB such as Freebase~\cite{freebase}. These approaches  usually achieve a high precision, but their scope is limited to the ontology of the knowledge graph. While KB QA is undoubtedly an important part of open-domain QA, we mainly discuss literature in text-based QA, which is most relevant to our work. 

Sentence-level QA has been studied since early 2000s, some of the most notable datasets being TrecQA~\cite{trecqa} and WikiQA~\cite{wikiqa}. See~\citet{prager2007open} for a comprehensive overview of early work.  With the advancement of deep neural networks and the availability of massive QA datasets such as SQuAD~\cite{squad}, open-domain phrase-level question answering has gained a great popularity~\cite{shen2017reasonet,raiman2017globally,min2018efficient,raison2018weaver,das2018multi}, where a few (5-10) documents relevant to the question are retrieved and then a deep neural model finds the answer in the document. 
Most previous work on open-domain QA has focused on mitigating error propagation of retriever models in a pipelined setting~\cite{chu2012finding}. For instance, retrieved documents could be re-ranked using reinforcement learning~\cite{wang2017r}, distant supervision~\cite{lin2018denoising}, or multi-task learning~\cite{nishida2018retrieve}. Several studies have also shown that answer aggregation modules could improve performance of the pipelined models~\cite{wang2017evidence,lee2018ranking}. 
% However, as we put more modules for document re-ranking or answer aggregation, efficiency of pipelined models inevitably becomes worse. In contrast to pipelining an expensive model after the document is retrieved, our method obtains the answer only through retrieval, which allows a highly efficient model with model transparency.

Our work is motivated by \citet{piqa} and adopts the concept and the advantage of using phrase index for large-scale question answering, though they only experiment in a close-domain (vanilla SQuAD) setup. 
% \hanna{Minjoon, people ask we need to have more description about differences in the modeling. Something like this: the previous work does not have scalar parameter, doesn't show how to index and encode phrases, and does not have a sparse vector, and our nbumber is much higher.  }

\paragraph{Approximate similarity search} Sublinear-time search for the nearest neighbor from a large collection of vectors is a significant interest to the information retrieval community~\cite{lsa,lda}.
In metric space (L1 or L2), one of the most classic search algorithms is Locality-Sensitive Hashing (LSH)~\cite{gionis1999similarity}, which uses a data-independent hashing function to map nearby vectors to the same cell. Stronger empirical performance has been observed with a data-dependent hashing function~\cite{andoni2015optimal} or k-means clustering for defining the cells. More recently, graph-based search algorithms~\cite{hnsw} have gained popularity as well. In non-metric space such as inner product, asymmetric Locality Sensitive Hashing (aLSH)~\cite{alsh2} is considered, where maximizing inner product search can be transformed into minimizing L2 distance by appending a single dimension to the vectors. While these methods are widely used for dense vectors, for extremely sparse data (such as document tf-idf with stop words), it is often more efficient to construct an inverted index and only look up items that have common hot dimensions with the query.

% \paragraph{Contextualized word representations} As demonstrated in Section~\ref{sec:model}, our model is a shallow layer constructed on top of a generic contextualized word representations. Traditionally, LSTM out of box has been considered as the standard de facto for encoding a sequence of word representations. More recently, a very large model trained on a large corpus of unlabeled data have shown significantly stronger performance. Most notably, ELMo~\cite{elmo} train a language model with several layers of LSTM on Wikipedia corpus, whereas BERT~\cite{bert} uses Transformer~\cite{transformer}, which consists of several layers of self-attention, to train a masked langauge model on Wikipedia Corpus as well as Book Corpus. Our model is based on BERT, which has shown better performance than other contextualization models.

\paragraph{Generative question answering} Mapping the phrases in a document to a common vector space to that of the questions can be viewed as an exhaustive enumeration of all possible questions that can be asked on the document in the vector space, but without a surface-form decoder. It is worth noting that generative question answering~\cite{lewis2019generative} has the opposite property; while it has a surface-form decoder by definition, it cannot easily enumerate a compact list of all possible semantically-unique questions.

\paragraph{Memory networks} One can view the phrase index as an external memory~\cite{weston2014memory,memnet} where the key is the phrase vector and the value is the corresponding answer phrase span. 

\section{Overview}\label{sec:scale}
In this section, we formally define ``open-domain question answering'' and provide an overview of our proposed model.

\subsection{Problem Definition}
In this paper, we are interested in the task of answering factoid questions from a large collection of web documents in real-time. This is often referred to as open-domain question answering (QA). %\footnote{The knowledge corpus could be structured such as Freebase~\cite{freebase} or Wikidata, which turns the problem into a \emph{semantic parsing} task.}
We formally formulate the task as follows. We are given a fixed set of (Wikipedia) documents $\xm^1, \dots , \xm^K$ (where $K$ is the number of documents, often on the order of millions), and each document $\xm^k$ has $N_k$ words, $\xm^k_1, \dots , \xm^k_{N_k}$. The task is to find the answer $\am$ to the question $\qm = \qm_1, \dots , \qm_S$. 
Then an open-domain QA model is a scoring function $F$ for each candidate phrase span $\xm^k_{i:j}$ such that $\am=\argmax_{k, i, j} F(\xm^k_{i:j}, \qm)$.

\paragraph{Scalability challenge} While the formulation is straightforward, argmax-ing over the entire corpus is computationally prohibitive, especially if $F$ is a complex neural model.
To avoid the computational bottleneck, previous open-domain QA models adopt pipeline-based methods; 
that is, as illustrated in Figure~\ref{fig:teaser} left, a fast retrieval-based model is used (e.g. tf-idf) to obtain a few relevant documents to the question,
and then a neural QA model is used to extract the exact answer from the documents~\cite{drqa}.
However, the method is not \emph{efficient} enough for real-time usage because the neural QA needs to re-encode all the documents for every new question, which is computationally expensive even with modern GPUs, and not suitable for low-latency applications.

\subsection{Encoding and Indexing Phrases}\label{subsec:index}
Motivated by~\citet{piqa}, our model encodes query-agnostic representations of text spans in Wikipedia offline and obtains the answer in real-time by performing nearest neighbor search at inference time. 
%  At training,  our  system  encodes  and indexes  span phrases  as  potential  answer  candidates  independent of the question to alleviate the need for re-encoding the documents for every query. At inference, the questions are encoded in the same vector space, and the answer is retrieved by performing maximum inner product search with the indexed phrases. 
%To alleviate the need for re-encoding the documents for every query, \citet{piqa} proposed to \emph{decompose} $F$ into $G$ and $H$ such that $F(\xm^k_{i:j}, \qm) = G(\xm^k_{i:j}) \cdot H(\qm)$, where $\cdot$ denotes inner product, so that the documents can be just encoded once (via $G$) and one only needs to encode the incoming question and perform maximum inner product search.
%While \citet{piqa} suggested the concept of pre-indexing every (constituent) text span in Wikipedia for open-domain QA, it only experimented on SQuAD. 
%The contribution of this paper is to extend the idea to web-scale by building a hierarchical index of documents and phrases, as demonstrated in the next subsection (Section~\ref{subsec:index}). 
We represent each phrase span in the corpus (Wikipedia) with a dense vector and a sparse vector. The dense vector is effective for encoding syntactic and semantic cues, while the sparse vector is good at encoding precise lexical information.
That is, the embedding of each span  $(i, j)$ in the document $\xm^k$ is represented with
\begin{equation}\label{eqn:x}
    \xb^k_{i:j} = [\db^k_{i:j}, \sbo^k_{i:j}] \in \RR^{d^\text{d} + d^\text{s}}
\end{equation}
where $\db^k_{i:j} \in \RR^{d^\text{d}}$ is the dense vector and $\sbo^k_{i:j} \in \RR^{d^\text{s}}$ is the sparse vector for span $(i,j)$ in the $k$-th document. Note that $d^\text{d} \ll d^\text{s}$. 
% \paragraph{Latency} The hierarchical index is very fast because it only needs to perform inner product (after documents and phrases are pre-encoded and the question is embedded), and only once per layer.
% In fact, while the search space is quite large, obtaining the exact answer is feasible if we have a sufficiently well-designed distributed system.\footnote{For instance, a GPU with 10 teraflops can (theoretically) process an inner product on a billion 1000D vectors in 0.1 second. Indexing the entire Wikipedia results in about 3 billion vectors, which might be even manageable with a single GPU, though keep in mind that data transfer speed will be a major bottleneck.} One can also consider approximate similarity search (See Section~\ref{sec:related}) or beam search that allows a low latency search even with a single CPU. 
% \paragraph{Homogeneity} It is worth noting that the model is transparent and homogeneous, and this allows us to easily develop or utilize fast inner product search algorithms or hardware (i.e. GPU) in a modular way.
This is also illustrated in Figure~\ref{fig:teaser} right.
Text span embeddings ($\xb^k_{i:j}$) for all possible $i, j, k$ pairs with $j - i < J$, where $J$ is maximum span length (i.e. all possible spans from all documents in Wikipedia), are pre-computed and stored as a \emph{phrase index}. Then at inference time, we embed each question into the same vector space, $\qb = [\db', \sbo'] \in \RR^{d^\text{d} + d^\text{s}}$. Finally, the answer to the question is obtained by finding the maximum inner product between $\qb$ and $\xb^k_{i:j}$,
\begin{equation}\label{eqn:answer}
k^*, i^*, j^* = \argmax_{k, i, j} \qb \cdot \xb^k_{i:j}.
\end{equation}
Needlessly to say, designing a good phrase representation model is crucial, which will be discussed in Section~\ref{sec:model}. Also, while inner product search is much more efficient than re-encoding documents, the search space is still quite large, such that exact search on the entire corpus is still undesirable. We discuss how we perform inner product search efficiently in Section~\ref{sec:training}.

\section{Phrase and Question Embedding}\label{sec:model}
In this section, we  first explain the embedding model for the dense vector in Section~\ref{subsec:dense}. Then we describe the embedding model for the sparse vector in Section~\ref{subsec:sparse}. Lastly, we describe the corresponding question embedding model to be queried on the phrase index in Section~\ref{subsec:question}.
For the brevity of the notations, we omit the superscript $k$ in this section since we do not learn cross-document relationships.

\subsection{Dense Model} \label{subsec:dense}

% In this subsection, we describe the model for obtaining each component of the phrase vector $\pb^k_{i:j} = [\sbo^k_i, \eb^k_j, c^k_{i,j}]$ for each answer candidate $(i, j)$ in the document $k$ (see Equation~\ref{eqn:x}).

The dense vector is responsible for encoding syntactic or semantic information of the phrase with respect to its context.
We decompose the dense vector $\db^k_{i:j}$ (Equation~\ref{eqn:x}) into three components: a vector that corresponds to the start position of the phrase, a vector that corresponds to the end position, and a scalar value that measures the coherency between the start and the end vectors. Representing phrases as a function of start and end vectors allows us to efficiently compute and store the vectors instead of enumerating all possible phrases (discussed in Section~\ref{subsec:dumping}).\footnote{Our phrase encoding is analogous to how  existing QA systems obtain the answer by predicting its start and the end positions. } 

The coherency scalar allows us to avoid non-constituent phrases during inference. For instance, consider a sentence such as ``Barack Obama was the 44th President of the US. He was also a lawyer.'' and when a question ``What was Barack Obama's job?'' is asked. Since both answers ``44th President of the US'' and ``lawyer'' are technically correct, we might end up with the answer that spans from ``44th'' to ``lawyer'' if we model start and end vectors independently. The coherency scalar helps us avoid this  by modeling it as a function of the start position and the end position. Formally, after phrase vector decomposition into dense and sparse, we can expand the dense vector into
\begin{equation}\label{eqn:x_x}
    \db_{i:j} = [\ab_i, \bb_j, c_{i,j}] \in \RR^{2d^\texttt{b} + 1}
\end{equation}
where $\ab_i, \bb_j \in \RR^{d^\texttt{b}}$ are the start and end vectors for the $i$-th and $j$-th words of the document, respectively; and $c_{i,j} \in \RR$ is the phrasal coherency scalar between $i$-th and $j$-th positions (hence $d^\texttt{d} = 2d^\texttt{b} + 1$).

To obtain these components of the dense vector, we leverage available contextualized word representations, in particular BERT-large~\cite{bert}, which is pretrained on a large corpus (Wikipedia and BookCorpus) and has proved to be very powerful in numerous natural language tasks.
% \paragraph{Basic Encoder} Following ~\citet{piqa}, the basic encoder first applies Bidirectional LSTM on the embeddings of the input words, and then use (single-head) self-attention layer on top of the LSTM outputs. The final output is the concatenation of the LSTM output and the self-attention output. We append a dummy token in the beginning (for consistency with BERT we use \texttt{[CLS]}) for the representation of the entire sequence.
BERT maps a sequence of the document tokens $\xm = \xm_1, \dots , \xm_N$ to a sequence of corresponding vectors (i.e. a matrix) $\Hb = [\hb_1; \dots ; \hb_N] \in \RR^{N \times d}$,  where $N$ is the length of the input sequence, $d$ is the hidden state size, and $[;]$ is vertical concatenation. We obtain the three components of the dense vector from these contextualized word representations.

We fine-tune BERT to learn a $d$-dimensional vector $\hb_i$ for encoding each token $\xm_i$. Every token encoding is split into four vectors $\hb_i=[\hb^1_i, \hb^2_i, {\hb^3_i},\hb^4_i] \in \mathbb{R}^d$, where $[,]$ is a column-wise concatenation.
Then we obtain the dense start vector $\ab_i$  from  $\hb^1_i$ and dense end vector   $\bb_j$ from  $\hb^2_j$. Lastly, we obtain the coherency scalar $c^k_{i,j}$ from the inner product of ${\hb^3_i}$ and $\hb^4_j$. The inner product allows more coherent phrases to have more similar start and end encodings.
That is, 
\begin{equation}\label{eqn:pij}
    \db_{i:j} = [\hb^1_i, \hb^2_j, {\hb^3_i} \cdot \hb^4_j] \in \RR^{2d^\texttt{b} + 1}
\end{equation}
where $\cdot$ indicates inner product operation and $\hb^1_i, \hb^2_j \in \mathbb{R}^{d^\texttt{b}}$ and  $\hb^3_i, \hb^4_j \in \mathbb{R}^{d^\texttt{c}}$ (hence $2d^\texttt{b} + 2d^\texttt{c} = d$). %Intuitively, the $\hb^1_i$ and $\hb^2_i$ encode if the token $i$ is a start or end of a phrase. 
%Intuitively, the inner product $\hb^3_i \cdot \hb^4_j$ allows more coherent phrases have more similar vector encodings for $\hb^3_i$ and $\hb^4_j$ splits of the token encoding.  \hanna{I had liked to mention that in our experiments we observe that this encoding has good results; Instead I thought we can say following piqa we did this}

%Specifically, to obtain $\ab_i, \bb_j, c_{i,j}$, we first split the (BERT) encoding into four matrices, $\Hb^1, \Hb^2 \in \RR^{N \times d^\texttt{b}}$ and $\Hb^3, \Hb^4 \in \RR^{N \times d^\texttt{c}}$ where $2d^\texttt{b} + 2d^\texttt{c} = d$ (hence $d^\texttt{d}$ depends on $d^\texttt{b}$ given $c$).
% That is, $\Hb = [\Hb^1, \Hb^2, \Hb^3, \Hb^4]$ where $[,]$ is column-wise concatenation.

% Then we obtain the dense start vector $\ab_i$ from the $i$-th column of $\Hb^1$ (i.e. $\ab_i = \hb^1$) and the dense end vector $\bb_j$ from the $j$-th column of $\Hb^2$ (i.e. $\bb_j = \hb^2$). Lastly, we obtain the coherency scalar $c^k_{i,j}$ from the inner product of $i$-th column of $\Hb^3$ and $j$-th column of $\Hb^4$.
% \ms{no matrix notation}

% \hanna{The rest of the sections are also written using matrices; you might want move it back to matrix notation, or define matrix afterwards}

\subsection{Sparse Model} \label{subsec:sparse}
We use term-frequency-based encoding to obtain the sparse embedding $\sbo^k_{i:j}$ for each phrase. 
Specifically, we largely follow DrQA~\cite{drqa} to construct 2-gram-based tf-idf, resulting in a highly sparse representation ($d^\texttt{d} \approx$16M) for each document. The sparse vectors are normalized so that the inner product effectively becomes cosine similarity.  We also compute a paragraph-level sparse vector in a similar way and add it to each document sparse vector for a higher sensitivity to local information. Note that, however, unlike DrQA where the sparse vector is merely used to retrieve a few (5-10) documents, we concatenate the sparse vector to the dense vector to form a standalone single phrase vector as in Equation~\ref{eqn:x}.
% \ms{put a footnote for trying out learned tfidf}

\subsection{Question Embedding Model}\label{subsec:question}
At inference, the question is encoded as  $\qb = [\db', \sbo'] = [\ab', \bb', c', \sbo']$ with the same number of components as the phrase index.
To obtain the dense query vector $\db' = [\ab', \bb', c']$, we use a special token (\texttt{[CLS]} for BERT) which is appended to the front of the question words (i.e. input question words are $\qm = \texttt{[CLS]}, \qm_1, \dots, \qm_S$). This allows us to model the dense query embedding differently from the dense embedding in the phrase index while sharing all parameters of the BERT encoder. That is, given the contextualized word representations of the question, we obtain the the dense query vector by 
\begin{equation}\label{eqn:question}
\pb' = [\hb'^1_1, \hb'^2_1, {\hb'^3_1} \cdot \hb'^4_1], 
\end{equation}
where $\hb'^1_1$ is the encoding corresponding to the (first) special token and we obtain the others in a similar way.
To obtain the sparse query vector $\sbo'$, we use the same tf-idf embedding model (Section ~\ref{subsec:sparse}) on the entire query.

\section{Training, Indexing \& Search}\label{sec:training}
Open-domain QA is a web-scale experiment, dealing with billions of words in Wikipedia while aiming for real-time inference. Hence (1) training the models, (2) indexing the embeddings, and (3) performing inner product search at inference time are non-trivial for both (a) computational time and (b) memory efficiency.
In particular, we carry out this section assuming that we have a constrained hardware environment of 4 P40 GPUs, 128 GB RAM, 16 cores and 2 TB of SATA SSD storage, to promote reproducibility of our experiments under academic setting.\footnote{Training takes 16 hours (64-GPU hours) and indexing takes 5 days (500 GPU-hours).}

\subsection{Training}\label{subsec:train}
As discussed in Section~\ref{subsec:sparse}, the sparse embedding model is trained in an unsupervised manner. For training the dense embedding model, instead of directly optimizing for Equation~\ref{eqn:answer} on entire Wikipedia, which is computationally prohibitive, we provide the golden paragraph to each question during training (i.e. SQuAD v1.1 setting).

Given the dense phrase and question embeddings, we first expand Equation~\ref{eqn:answer} by substituting Equation~\ref{eqn:pij} and Equation~\ref{eqn:question} (omitting document terms):
\begin{align*}
    i^*, j^* & =  \argmax_{i,j} \db' \cdot \db_{i:j} \\
             & = \argmax_{i,j} \hb'^1_1 \cdot \hb^1_i + \hb'^2_1 \cdot \hb^2_j + \\
             & \hb'^3_1 \cdot \hb'^4_1 + \hb^3_i \cdot \hb'^4_j 
\end{align*}
% For the brevity of the notations,
From now on we let $l^1_i = \hb'^1_1 \cdot \hb^1_i$ (phrase start logits), $l^2_j = \hb'^2_1 \cdot \hb^2_j$ (phrase end logits), and $l_{i,j} = l^1_i + l^2_j + \hb'^3_1 \cdot \hb'^4_1 + \hb^3_i \cdot \hb'^4_j$ i.e. the value that is being maximized in the above equation.

One straightforward way to define the loss is to define it as the negative log probability of the correct answer where $\paramp(i,j) \propto \exp({l_{i,j}})$. In other words,
\begin{equation}\label{eqn:true-loss}
    L=-l_{i^*,j^*} + \log \sum_{i,j} \exp(l_{i,j})
\end{equation}
where $L$ is the loss to minimize.
% However, since there are so many terms in the summation ($O(T^2)$ terms), the log summation is much larger than $l_{i^*,j^*}$ and does not sensitively react to changes in individual terms' values, which results in small gradient.
% To encourage a higher gradient, we obtain $\Lb^1 = $
% \begin{equation}
%     L^1 =  + \log \sum_i \exp{l^1_i} + \log \sum_j \exp{l^2_j}.
% \end{equation}
% each log summation has only $O(T)$ terms, so they are much more sensitive.
Note that explicitly enumerating all possible phrases (enumerating all $(i,j)$ pairs) during training time would be memory-intensive.
Instead, we can efficiently obtain the loss by:
\begin{align*}
\lb^1 & = [l^1_1, \dots, l^1_T] = \qb^1 {\Hb^1}^\top \\
\lb^2 & = \hb^2_1 {\Hb^2}^\top\\
\Lb & = \Hb^3 {\Hb^4}^\top + {\lb^1}^\top + \lb^2 
\end{align*}
where $\Hb^m = [\hb^m_1, \dots, \hb^m_T]$ for $m=1,2,3,4$,  $+$ is with broadcasting and $(i,j)$-th element of $\Lb$ is $l_{i,j}$. Note that $L$ can be entirely computed from $\Lb$.

While the loss function is clearly unbiased with respect to $\paramp(i,j) \propto \exp({l_{i,j}})$, the summation in Equation~\ref{eqn:true-loss} is computed over $T^2$ terms which is quite large and causes small gradient. 
To aid training, we define an auxilary loss $L^1$ corresponding to the start logits,
\begin{equation}
    L^1 =  -l^1_{i^*} + \log \sum_i \exp(\frac{1}{T}\sum_j l_{i,j})
\end{equation}
and $L^2$ for the end logits in a similar way. By early summation (taking the mean), we reduce the number of exponential terms and allow larger gradients. We average between the true and aux loss for the final loss: $\frac{L}{2} + \frac{L^1 + L^2}{4}$. 

\paragraph{No-Answer Bias} During training SQuAD (v1.1), we never observe negative examples (i.e. an unanswerable question in the paragraph). Following~\citet{levy2017zero}, we introduce a trainable no-answer bias when computing softmax. For each paragraph, we create two negative examples by bringing one question from another article and one question from the same article but different paragraphs. Instead of randomly sampling, we bring the question with the highest inner product (i.e. most similar) with a randomly-picked positive question in the current paragraph, using a question embedding model trained on SQuAD v1.1.
We jointly train the positive examples with the negative examples. 

% We only use the augmented dataset for open-domain setup (Section~\ref{subsec:open}) and not for vanilla SQuAD (Section~\ref{subsec:close}).

\subsection{Indexing}\label{subsec:dumping}

Wikipedia consists of approximately 3 billion tokens, so enumerating all phrases with length $\leq 20$ will result in about 60 billion phrases. With 961D of \texttt{float32} per phrase, one needs 240 TB of storage (60 billion times 961 dimensions times 4 bytes per dimension). While not impossible in industry scale, the size is clearly out of reach for independent or academic researchers and critically unfriendly for open research.
We discuss three techniques we employee to reduce the size of the index to 1.2 TB without sacrificing much accuracy, which becomes much more manageable for everyone.
In practice, additional 300-500GB will be needed to store auxiliary information for efficient indexing, which still sums up to less than 2TB.

\paragraph{1. Pointer} Since each phrase vector is the concatenation of $\ab_i$ and  $\bb_j$ (and a scalar $c_{i,j}$ but it takes very little space), many phrases share the same start or end vectors. Hence we store a single list of the start and the end vectors independently and just store pointers to those vectors for the phrase representation. This effectively reduces the memory footprint from 240 TB to 12 TB.

\paragraph{2. Filtering} We train a simple single-layer binary classifier on top of each of the start and end vectors, supervised with the actual answer (without observing the question). This allows us to not store vectors that are unlikely to be a potential start or end position of the answer phrase, further reducing the memory footprint from 12 TB to 5 TB.

\paragraph{3. Quantization} 
We reduce the size of each vector by scalar quantization (SQ). That is, we convert each \texttt{float32} value to \texttt{int8} with appropriate offset and scaling. This allows us to reduce the size by one-fourth. Hence the final memory consumption is 1.2 TB.
In future, more advanced methods such as Product Quantization (PQ)~\cite{pq} can be considered, though we note that in our experiment setup we could not find a good configuration that does not drop the accuracy significantly.

\subsection{Search}\label{subsec:search}
% As we discussed in Section~\ref{sec:related}, many maximum inner product search (MIPS) techniques for dense or sparse vectors exist for sublinear time approximation, which allows us to directly approximate the argmax in Equation~\ref{eqn:answer}. 
While it would be ideal to (and possible to) directly approximate argmax in Equation~\ref{eqn:answer} by using sparse maximum inner product search algorithm (some discussed in Section~\ref{sec:related}), we could not find a good open-source implementation that can scale up to billions of vectors and handle the dense and the sparse part of the phrase vector at the same time. We instead consider three approximation strategies. 

First, \emph{sparse-first search} (SFS) approximates the argmax by retrieving top-$k^\text{s}$ documents with the sparse similarity search and then performing exact search (including sparse inner product scores) over all the phrases in retrieved documents.
This is analogous to most pipeline-based QA systems, although our model can still yield much higher speed because it only needs to perform inner product once the documents are retrieved.
Since the number of sparse document vectors is relatively small (5 million), we directly perform exact search using \texttt{scipy}, which has under 0.2s latency per query.

Second, \emph{dense-first search} (DFS) approximates the argmax by doing search on the dense part first to retrieve top-$k^\text{d}$ vectors and then reranking them by accessing the corresponding sparse vectors. Note that this implies a widely different behavior from SFS, as described in Section~\ref{subsec:open}.
We use \texttt{faiss}~\cite{faiss}, open-sourced and large-scale-friendly similarity search package for dense vectors.

Lastly, we consider a \emph{hybrid} approach by independently performing both search strategies and reranking the appended list of the results.

Also, instead of directly searching on the dense vector $\db_{i:j}$ (concatenation of start, end, and coherency), we first search on the start vector $\ab_i$ and obtain the best end position for each retrieved start position by computing the rest. We found that this allows us to save memory and time without sacrificing much accuracy, since the start vectors alone seem to contain sufficiently rich syntactic and semantic information already that makes the search possible even in a large scale.

% In this work, however, we choose to query on the document index first to retrieve a few documents and then perform (exact) search on the (pre-computed) phrase indexes corresponding to the retrieved documents to obtain the final answer. Our decision is largely based on following reasons. First, little open-sourced MIPS solutions exist for a combination of sparse and dense vectors, and one probably has to reach proprietary software or build one's own. Second, as~\citet{drqa} has shown, about 80\% of questions in open-domain SQuAD are answerable by top-5 document retrieval, so global approximation is not yet a bottleneck given the current state-of-the-art performance (below 40\% EM). Third, this allows us to have an apple-to-apple comparison with pipelined systems which also only look at top-k documents. We would like to also mention that since we are performing exact search, our latency measure should be considered as the lower bound, and further speed up can be easily expected with a dedicated search index.

\section{Experiments}\label{sec:exp}
\begin{table}[]
    \centering
    \resizebox{\columnwidth}{!}{%
    \begin{tabular}{p{1.5cm}lccc}
        \toprule
                      & Model   & EM    & F1 & W/s \\
        \midrule
         
        \multirow{2}{2cm}{Original} &  DrQA  & 69.5 & 78.8 & 4.8K\\
        & BERT-Large  & 84.1 & 90.9 & 51 \\
        \midrule
         
        \multirow{3}{2cm}{Query-Agnostic} & LSTM+SA           & 49.0 & 59.8 & -\\
        & LSTM+SA+ELMo      & 52.7 & 62.7 & -\\
        % Base \ours* (127D)      & 65.9 & 74.7  \\
        % PI-BERT Base 511D  & 78.9  & 70.7  \\
        % PI-BERT Base 255D  & 76.2  & 68.1  \\
        % PI-BERT Base 127D  & 74.2  & 66.2  \\
        & \ours\ (dense only)      & 73.6 & 81.7 & 28.7M \\
        & $+$ Linear layer & 66.9 & 76.4 & - \\
        & $+$ Indep. encoders & 65.4 & 75.1 & - \\
        & $-$ Coherency scalar       & 71.5 & 81.5 & - \\
        % Large \ours* (127D)     & 67.1 & 75.4 \\
        % PI-BERT Large 511D &       &  \\
        % PI-BERT Large 255D & 78.1  & 70.1 \\
        % PI-BERT Large 127D & 77.0  & 69.2 \\
        \bottomrule
    \end{tabular} 
    }
    \caption{Results on SQuAD v1.1. `W/s' indicates number of words the model can process (read) per second on a CPU in a batch mode (multiple queries at a time). DrQA~\cite{drqa} and BERT~\cite{bert} are from SQuAD leaderboard, and LSTM+SA and LSTM+SA+ELMo are  query-agnostic baselines from~\citet{piqa}.}
    \label{tab:pi-squad}
\end{table}

Experiment section is divided into two parts. 
First, we report results on SQuAD~\cite{squad}. This can be considered as a small-scale prerequisite to the open-domain experiment. It also allows a convenient comparison to state-of-the-art models in SQuAD, especially on the speed of the model. Under a fully controlled environment and batch-query scenario, our model processes words nearly 6,000 times faster than DrQA~\cite{drqa}.
Second, we report results on open-domain SQuAD (called SQuAD-Open), following the same setup as in DrQA. We show that our model achieves up to 6.4\% better accuracy and up to 58 times faster end-to-end inference time than DrQA while exploring nearly 200 times more unique documents.
All experiments are CPU-only benchmark.

\subsection{SQuAD v1.1 Experiments}\label{subsec:close}
In the SQuAD v1.1 setup, our model effectively uses only the dense vector since every sparse (document tf-idf) vector will be identical in the same paragraph. While this is a much easier problem than open-domain, it can serve as a reliable and fast indicator of how well the model would do in the open-domain setup.

\paragraph{Model details} We use BERT-large ($d=1024$) for the text encoders, which is pretrained on a large text corpus (Wikipedia dump and Book Corpus). We refer readers to the original paper by~\citet{bert} for details; we mostly use the default settings described there. We use $d^\texttt{b} = 480$, resulting in phrase size of $2d^\texttt{b} + 1 = 961$, and  $d^\texttt{c} = 32$. We train with a batch size of 12 (on four P40 GPUs) for 3 epochs. 

% \paragraph{Comparisons}
% \hanna{move the comparison models here and in results just discuss the results and analysis}
% \hanna{name the comparison models well and use the same name throughout the text and the tables.}

\paragraph{Baselines} We compare the performance of our system~\ours\ with a few baselines in terms of accuracy and efficiency. The first group are among the models that are submitted to SQuAD v1.1 Leaderboard, specifically DrQA~\cite{drqa} and BERT~\cite{bert} (current state of the art).  These models  encode the evidence document given the question, but they  suffer  from  the  disadvantage  that  the  evidence document  needs  to  be  re-encoded  for every new question at the inference time, and they  are  strictly  linear time in that they cannot utilize approximate search algorithms. The second group of baselines are introduced by \citet{piqa}, specifically LSTM+SA and LSTM+SA+ELMo that also encode phrases independent of the question using LSTM, Self-Attention, and ELMo~\cite{elmo} encodings. 

\paragraph{Results} Table~\ref{tab:pi-squad} compares the performance of our system with different baselines in terms of efficiency and accuracy.
We note the following observations from the result table. (1) \ours\ outperforms the query-agnostic baseline~\cite{piqa} by a large margin, 20.1\% EM and 18.5\% F1. This is largely credited towards the usage of BERT encoder with an effective phrase embedding mechanism on the top. (2) \ours\ outperforms DrQA by 3.3\% EM. This signifies that phrase-indexed models can now outperform early (unconstrained) state-of-the-art models in SQuAD. (3) \ours\ is 9.2\% below the current state of the art. The difference, which we call \emph{decomposability gap}\footnote{The gap is due to constraining the scoring function to be decomposable into question encoder and context encoder.}, is now within 10\% and future work will involve further closing the gap. (4) Query-agnostic models can process (read) words much faster than query-dependent representation models. In a controlled environment where all information is in memory and the documents are pre-indexed, \ours\ can process 28.7 million words per second, which is 6,000 times faster than DrQA and 563,000 times faster than BERT without any approximation.

\paragraph{Ablations} Ablations are also shown at the bottom of Table~\ref{tab:pi-squad}.
The first ablation adds a linear layer on top of the BERT encoder for the phrase embeddings, which is more analogous to how BERT handles other language tasks. We see a huge drop in performance. We also try independent BERT encoders (i.e. unshared parameters) between phrase and question embedding models, and we also see a large drop as well. These seem to indicate that a careful design consideration for even small details are crucial when finetuning BERT.    
Our ablation that excludes coherency scalar decreases \ours's EM score by 2\% and F1 by 0.2\%. This agrees with our intuition that the coherency scalar is useful for precisely defining valid phrase constituents.
% \paragraph{Model ablations.} What happens if you add things and subtract things? Discuss them here. See Figure~\ref{tab:ab}. \hanna{An important ablation is to understand how the coherency scalar for phrase reprsentation helped here?}

% \input{05-exp-f2}
% \input{06-exp-t2}

\begin{table}[]
    \centering
    \resizebox{0.88\columnwidth}{!}{%
    \begin{tabular}{lcccccc}
        \toprule
                      & F1    & EM & s/Q & \#D/Q   \\
        \midrule
        DrQA          & -  & 29.8 & 35 & 5 \\
        R$^3$          & 37.5 & - & - & - \\
        Paragraph ranker      & -  & 30.2 & - & 20 \\
        Multi-step reasoner      & 39.2 & 31.9 & - & - \\
        MINIMAL      & 42.5 & 34.7 & - & 10 \\
        BERTserini & 46.1 & 38.6 & 115 & - \\
        Weaver & - & 42.3 & - & 25 \\
        % 218, 5298 in batch mode
        \midrule
        % LSTM+SA+ELMo & 19.0 &  13.2 & 8500 & 100000  \\
        % Base PI-BERT* (127D)  & X  & X & X & X  \\
        \ours-SFS &  42.5 & 33.3 & 0.60 & 5\\
        \ours-DFS &  35.9 & 28.5 & 0.51 & 815\\
        --sparse scale=0 &  16.3 & 11.2 & 0.40 & 815\\
        \ours-Hybrid &  44.4 & 36.2 & 0.81 & 817\\
        % Large PI-BERT* (127D) & X  & X & X & X \\
        \bottomrule
        
    \end{tabular}
    }
    \caption{Results on SQuAD-Open. Top rows are previous models that re-encode documents for every question.
    The bottom rows are our proposed model. `s/Q' is seconds per query on a CPU and `\#D/Q' is the number of documents visited per query.}
    \label{tab:od-squad}
\end{table}

\subsection{Open-domain Experiments}\label{subsec:open}
In this subsection, we evaluate our model's performance (accuracy and speed) on Open-domain SQuAD (SQuAD-Open), which is an extension of SQuAD~\cite{squad} by~\citet{drqa}. In this setup, the evidence is the entire English Wikipedia, and the golden paragraphs are not provided for questions.  
%While Section~\ref{subsec:close} provides a convenient benchmark of our proposed model's performance, what we are really interested in is its performance in open-domain setup. In particular, we are not only interested in the accuracy (EM and F1), but its trade-off with the latency per query.
% In this section, we compare our model with other previous models in Open-domain SQuAD, which is an extension of SQuAD by~\citet{drqa}. In this setup, the evidence is the entire Wikipedia, and the golden paragraphs are not provided for questions. % where the system is given the entire Wikipedia corpus and the questions are asked without the golden paragraphs that they belong to.

% \input{06-exp-f2}

\paragraph{Model details} For the dense vector, we adopt the same setup from Section~\ref{subsec:close} except that we train with no-answer questions (Section~\ref{subsec:train}) and an increased batch size of 18.  For the sparse vector of each phrase, we use the identical 2-gram tf-idf vector used by~\citet{drqa}, whose vocabulary size is approximately 17 million, of the document that contains the phrase. Since the sparse vector and the dense vector are independently obtained, we tune the linear scale between the sparse and the dense vectors and found that 0.05 (multiplied on the sparse vector) gives the best performance. As discussed in Section~\ref{subsec:search}, we consider three search strategies. For sparse-first search (SFS), we retrieve top-5 documents.
For dense-first search (DFS), we use an HNSW-based~\cite{hnsw} coarse quantizer with $2^{20}$ (1M) clusters (obtained with k-means) and nprobe=64 (number of clusters to visit). We retrieve top 1000 dense (start) vectors. The `Hybrid' setup adopts the same configurations from both.

\paragraph{Baselines} We compare our system with previous state-of-the-art models for open-domain question answering. The baselines include DrQA~\cite{drqa}, MINIMAL~\cite{min2018efficient}, multi-step-reasoner~\cite{das2018multi}, Paragraph Ranker~\cite{lee2018ranking}, $R^3$~\cite{wang2017r}, BERTserini~\cite{yang2019end}, and Weaver~\cite{raison2018weaver}.
We do not experiment with~\citet{piqa} due to its poor performance with respect to~\ours\ as demonstrated in Table~\ref{tab:pi-squad}.

\begin{table}[]
    \centering
    \resizebox{\columnwidth}{!}{%
    \begin{tabular}{ll}
        \toprule
        \multicolumn{2}{l}{Q: What can hurt a teacher's mental and physical health?}\\
        \multicolumn{2}{l}{A: occupational stress} \\
        \midrule
        DrQA & [\textit{Mental health}] ... and poor mental health can lead \\
        & to problems such as \textbf{substance abuse}. \\
        \ours & [\textit{Teacher}] Teachers face several occupational hazards \\
        & in their line of work, including \textbf{occupational stress}, ... \\
        
        \midrule
        \multicolumn{2}{l}{Q: Who was Kennedy's science adviser that opposed manned} \\
        \multicolumn{2}{l}{spacecraft flights?} \\
        \multicolumn{2}{l}{A: Jerome Wiesner} \\
        \midrule
        DrQA & [\textit{Apollo program}] Kennedy's science advisor \textbf{Jerome} \\
        & \textbf{Wiesner}, (...) his opposition to manned spaceflight ...\\
        & [\textit{Apollo program}] ... and the sun by NASA manager \\
        & \textbf{Abe Silverstein}, who later said that ...\\
        & [\textit{Apollo program}] Although Grumman wanted a second \\ 
        & unmanned test, \textbf{George Low} decided (...) be manned. \\
        \ours & [\textit{Apollo program}] Kennedy's science advisor \textbf{Jerome} \\
        & \textbf{Wiesner}, ... his opposition to manned spaceflight ...\\
        & [\textit{Space Race}] \textbf{Jerome Wiesner} of MIT, who served as a \\
        & (...) advisor to (...) Kennedy, (...) opponent of manned ...\\
        & [\textit{John F. Kennedy}] ... science advisor \textbf{Jerome Wiesner} \\
        & (...) strongly opposed to manned space exploration, ... \\
        
        \midrule
        \multicolumn{2}{l}{Q: What to do when you're bored?} \\
        % \multicolumn{2}{l}{A: Jerome Wiesner} \\
        \midrule
        DrQA & [\textit{Bored to Death (song)}] I'm nearly bored to \textbf{death} \\
        & [\textit{Waterview Connection}] The twin tunnels were bored \\
        & by (...) tunnel \textbf{boring} machine (TBM) ...\\
        & [\textit{Bored to Death (song)}] It's easier to say you're bored, \\
        & or to be angry, than it is to be \textbf{sad}. \\
        \ours & [\textit{Big Brother 2}] When bored, she enjoys \textbf{drawing}. \\
        & [\textit{Angry Kid}] Angry Kid is (...) bored of long car journeys, \\
        & so Dad suggests he just \textbf{close his eyes and sleep}.\\
        & [\textit{Pearls Before Swine}] In law school, he became so \\ 
        & bored during classes, he started to \textbf{doodle a rat}, ... \\
        \bottomrule
        
    \end{tabular}
    }
    \caption{Prediction samples from DrQA and \ours\ in open-domain (English Wikipedia). Each sample shows [\textit{document title}], context, and \textbf{predicted answer}.}
    \label{tab:od-samples}
\end{table}
\paragraph{Results} Table~\ref{tab:od-squad} shows the results of our system and previous models on SQuAD-Open. 
% We report both the accuracy (F1 and EM), inference speed (seconds per query, s/Q) of the models. We also Note that previous models  process the document on the fly, while our models load stored phrase indexes perform maximum inner product search. Our model explores two search strategies: sparse-first that first searchers through the sparse vector, and dense-first that first searches through the dense vector.  
We note following observations:
(1) \ours-Hybrid outperforms DrQA by 6.4\% while achieving 43 times faster inference speed.
% \footnote{We previously reported 6K times faster speed in Section~\ref{subsec:close}; there is a significant difference largely because \ours\ is covering a larger number of documents than DrQA and we need to account for the overhead during similarity search and disk access.}
(2) \ours-Hybrid is 6.1\% EM behind Weaver, which co-encodes top 25 documents (retrieved by tf-idf) for every new question. As mentioned in Section~\ref{subsec:close}, the difference between ours and Weaver can be considered as the \emph{decomposability gap} arising from the constraint of query-agnostic phrase representations. We note, however, that the gap is smaller now in open-domain, and the speed-up is expected to be much larger\footnote{Weaver is not open-sourced so we could not benchmark it.} since Weaver has higher computational complexity than DrQA and reads top 25 documents.
(3) We also report the number of documents that our model computes exact search on and compare it to that of DrQA, as indicated by `\#D/Q' in the table. Top-1000 dense search in \ours-Hybrid results in 817 unique documents on average, which is much more diverse than the 5 documents that DrQA considers. The benefit of this diversity is better illustrated in the upcoming qualitative analysis (Table~\ref{tab:od-samples}).

\paragraph{Ablations} Table~\ref{tab:od-squad} (bottom) shows the effect of different search strategies (SFS vs DFS vs Hybrid) and the importance of the sparse vector.

\emph{SFS vs DFS vs Hybrid}: We first see that \ours-SFS and \ours-DFS have comparable inference speed while \ours-SFS has 6.6\% higher F1. While this demonstrates the effectiveness of sparse search, it is important to note that this might be due to the high word overlap between the question and the context in SQuAD. Furthermore, we see that Hybrid achieves the highest accuracy in both F1 and EM, implying that the two strategies are complimentary.

\emph{Sparse vector}: \ours-DFS with `sparse scale=0' implies that we entirely remove the sparse vector, i.e. ${\xb_{i:j}} = {\db_{i:j}}$ in Equation~\ref{eqn:x}. While this wouldn't have any effect in SQuAD v.1.1, we see a significant drop (-19.6\% F1), indicating the importance of the sparse vector in open-domain for distinguishing semantically close but lexically distinct entities.

\begin{table}[]
    \centering
    \resizebox{\columnwidth}{!}{%
    \begin{tabular}{ll}
        \toprule
        \multicolumn{2}{l}{Q: What was the main radio network in the 1940s in America?}\\
        \multicolumn{2}{l}{A: NBC Red Network} \\
        \midrule
        \ours & [\textit{American Broadcasting Company}] In the 1930s, radio in \\
        & the United States was dominated by (...): the \textbf{Columbia}\\
        & \textbf{Broadcasting System}, the Mutual Broadcasting (...). \\
        
        \midrule
        
        \multicolumn{2}{l}{Q: Which city is the fifth-largest city in California?}\\
        \multicolumn{2}{l}{A: Fresno} \\
        \midrule
        \ours & [\textit{Oakland, California}] \textbf{Oakland} is the largest city \\
        & and the county seat of (...), California, United States.  \\
        
        \bottomrule
        
    \end{tabular}
    }
    \caption{Wrong prediction samples from \ours\ in open-domain (English Wikipedia). Each sample shows [\textit{document title}], context, and \textbf{predicted answer}.}
    \label{tab:error-samples}
\end{table}

\paragraph{Qualitative Analysis} Table~\ref{tab:od-samples} (and Table~\ref{tab:od-more-samples} in Appendix~\ref{sec:sample}) contrasts between the results from DrQA and \ours-Hybrid. In the top example, we note that DrQA fails to retrieve the right document, whereas \ours\ finds the correct answer. This happens exactly because the document retrieval model would not precisely know what kind of content is in the document, while dense search allows it to consider the content directly through phrase-level retrieval. In the second example, while both obtain the correct top-1, \ours\ also obtains the same answer from three different documents.  The last example (not from SQuAD) does not have a noun entity, in which a term-frequency-based search engine often performs poorly. We indeed see that DrQA fails because wrong documents are retrieved. On the other hand, \ours~is able to obtain good answers from several different documents. These results also reinforce the importance of exploring diverse documents (`\#D/Q' in Table~\ref{tab:od-squad}).

\paragraph{Error Analysis} %\jl{error analysis goes here.}
Table~\ref{tab:error-samples} shows wrong predictions from \ours. In the first example, the model seems to fail to distinguish `1940s' from `1930s'. In the second example, the model seems to focus more on the word `largest' than the word `fifth-' in the question.

% \subsection{Analysis}
% Some analyses go here.

\section{Conclusion}
We introduce a model for real-time open-domain question answering by learning indexable phrase representations independent of the query, which leverage both dense and sparse vectors to capture lexical, semantic, and syntactic information. 
% Our model is able to index all phrases in Wikipedia using efficient encoding of phrases with pointers to the start and end tokens.  Further, our model can retrieve an indexed phrase with maximum inner product search to answer a question.  
On SQuAD-Open, our experiments show that our model can read words 6,000 times faster under a controlled environment and 43 times faster in a real setup than DrQA while achieving 6.4\% higher EM. We believe that even further speedup and larger coverage of documents can be done with a dedicated similarity search package for dense+sparse vectors. 
We note that, however, the gap due to query-agnostic constraint still exists and is at least 6.1\% EM. Hence, more effort on designing a better phrase representation model is needed to close the gap.

\section*{Acknowledgement}
This research was supported by ONR (N00014-18-1-2826, N00014-17-S-B001), NSF (IIS 1616112), Allen Distinguished Investigator Award, Samsung GRO,  National Research Foundation of Korea (NRF-2017R1A2A1A17069645), and gifts from Allen Institute for AI, Google, and Amazon. We thank the members of UW NLP, Google AI, and the anonymous reviewers for their insightful comments.

\bibliography{00-main}

\begin{thebibliography}{35}
\expandafter\ifx\csname natexlab\endcsname\relax\def\natexlab#1{#1}\fi

\bibitem[{Andoni and Razenshteyn(2015)}]{andoni2015optimal}
Alexandr Andoni and Ilya Razenshteyn. 2015.
\newblock Optimal data-dependent hashing for approximate near neighbors.
\newblock In \emph{Proceedings of the forty-seventh annual ACM symposium on
  Theory of computing}.

\bibitem[{Berant et~al.(2013)Berant, Chou, Frostig, and Liang}]{kgqa}
Jonathan Berant, Andrew Chou, Roy Frostig, and Percy Liang. 2013.
\newblock Semantic parsing on freebase from question-answer pairs.
\newblock In \emph{EMNLP}.

\bibitem[{Blei et~al.(2003)Blei, Ng, and Jordan}]{lda}
David~M Blei, Andrew~Y Ng, and Michael~I Jordan. 2003.
\newblock Latent dirichlet allocation.
\newblock \emph{JMLR}.

\bibitem[{Bollacker et~al.(2008)Bollacker, Evans, Paritosh, Sturge, and
  Taylor}]{freebase}
Kurt Bollacker, Colin Evans, Praveen Paritosh, Tim Sturge, and Jamie Taylor.
  2008.
\newblock Freebase: a collaboratively created graph database for structuring
  human knowledge.
\newblock In \emph{SIGMOD}.

\bibitem[{Chen et~al.(2017)Chen, Fisch, Weston, and Bordes}]{drqa}
Danqi Chen, Adam Fisch, Jason Weston, and Antoine Bordes. 2017.
\newblock Reading wikipedia to answer open-domain questions.
\newblock In \emph{ACL}.

\bibitem[{Chu-Carroll et~al.(2012)Chu-Carroll, Fan, Boguraev, Carmel,
  Sheinwald, and Welty}]{chu2012finding}
Jennifer Chu-Carroll, James Fan, BK~Boguraev, David Carmel, Dafna Sheinwald,
  and Chris Welty. 2012.
\newblock Finding needles in the haystack: Search and candidate generation.
\newblock \emph{IBM Journal of Research and Development}.

\bibitem[{Das et~al.(2019)Das, Dhuliawala, Zaheer, and McCallum}]{das2018multi}
Rajarshi Das, Shehzaad Dhuliawala, Manzil Zaheer, and Andrew McCallum. 2019.
\newblock Multi-step retriever-reader interaction for scalable open-domain
  question answering.
\newblock In \emph{ICLR}.

\bibitem[{Deerwester et~al.(1990)Deerwester, Dumais, Furnas, Landauer, and
  Harshman}]{lsa}
Scott Deerwester, Susan~T Dumais, George~W Furnas, Thomas~K Landauer, and
  Richard Harshman. 1990.
\newblock Indexing by latent semantic analysis.
\newblock \emph{Journal of the American society for information science}.

\bibitem[{Devlin et~al.(2019)Devlin, Chang, Lee, and Toutanova}]{bert}
Jacob Devlin, Ming-Wei Chang, Kenton Lee, and Kristina Toutanova. 2019.
\newblock Bert: Pre-training of deep bidirectional transformers for language
  understanding.
\newblock In \emph{NAACL-HLT}.

\bibitem[{Gionis et~al.(1999)Gionis, Indyk, Motwani
  et~al.}]{gionis1999similarity}
Aristides Gionis, Piotr Indyk, Rajeev Motwani, et~al. 1999.
\newblock Similarity search in high dimensions via hashing.
\newblock In \emph{VLDB}.

\bibitem[{Jegou et~al.(2011)Jegou, Douze, and Schmid}]{pq}
Herve Jegou, Matthijs Douze, and Cordelia Schmid. 2011.
\newblock Product quantization for nearest neighbor search.
\newblock \emph{TPAMI}.

\bibitem[{Johnson et~al.(2017)Johnson, Douze, and J{\'e}gou}]{faiss}
Jeff Johnson, Matthijs Douze, and Herv{\'e} J{\'e}gou. 2017.
\newblock Billion-scale similarity search with gpus.
\newblock \emph{arXiv preprint arXiv:1702.08734}.

\bibitem[{Lee et~al.(2018)Lee, Yun, Kim, Ko, and Kang}]{lee2018ranking}
Jinhyuk Lee, Seongjun Yun, Hyunjae Kim, Miyoung Ko, and Jaewoo Kang. 2018.
\newblock Ranking paragraphs for improving answer recall in open-domain
  question answering.
\newblock In \emph{EMNLP}.

\bibitem[{Levy et~al.(2017)Levy, Seo, Choi, and Zettlemoyer}]{levy2017zero}
Omer Levy, Minjoon Seo, Eunsol Choi, and Luke Zettlemoyer. 2017.
\newblock Zero-shot relation extraction via reading comprehension.
\newblock In \emph{CoNLL}.

\bibitem[{Lewis and Fan(2019)}]{lewis2019generative}
Mike Lewis and Angela Fan. 2019.
\newblock Generative question answering: Learning to answer the whole question.
\newblock In \emph{ICLR}.

\bibitem[{Lin et~al.(2018)Lin, Ji, Liu, and Sun}]{lin2018denoising}
Yankai Lin, Haozhe Ji, Zhiyuan Liu, and Maosong Sun. 2018.
\newblock Denoising distantly supervised open-domain question answering.
\newblock In \emph{ACL}.

\bibitem[{Malkov and Yashunin(2018)}]{hnsw}
Yury~A Malkov and Dmitry~A Yashunin. 2018.
\newblock Efficient and robust approximate nearest neighbor search using
  hierarchical navigable small world graphs.
\newblock \emph{TPAMI}.

\bibitem[{Miller et~al.(2016)Miller, Fisch, Dodge, Karimi, Bordes, and
  Weston}]{memnet}
Alexander Miller, Adam Fisch, Jesse Dodge, Amir-Hossein Karimi, Antoine Bordes,
  and Jason Weston. 2016.
\newblock Key-value memory networks for directly reading documents.
\newblock \emph{arXiv preprint arXiv:1606.03126}.

\bibitem[{Min et~al.(2018)Min, Zhong, Socher, and Xiong}]{min2018efficient}
Sewon Min, Victor Zhong, Richard Socher, and Caiming Xiong. 2018.
\newblock Efficient and robust question answering from minimal context over
  documents.
\newblock In \emph{ACL}.

\bibitem[{Nishida et~al.(2018)Nishida, Saito, Otsuka, Asano, and
  Tomita}]{nishida2018retrieve}
Kyosuke Nishida, Itsumi Saito, Atsushi Otsuka, Hisako Asano, and Junji Tomita.
  2018.
\newblock Retrieve-and-read: Multi-task learning of information retrieval and
  reading comprehension.
\newblock In \emph{CIKM}.

\bibitem[{Peters et~al.(2018)Peters, Neumann, Iyyer, Gardner, Clark, Lee, and
  Zettlemoyer}]{elmo}
Matthew Peters, Mark Neumann, Mohit Iyyer, Matt Gardner, Christopher Clark,
  Kenton Lee, and Luke Zettlemoyer. 2018.
\newblock Deep contextualized word representations.
\newblock In \emph{NAACL-HLT}.

\bibitem[{Prager et~al.(2007)}]{prager2007open}
John Prager et~al. 2007.
\newblock Open-domain question--answering.
\newblock \emph{Foundations and Trends{\textregistered} in Information
  Retrieval}.

\bibitem[{Raiman and Miller(2017)}]{raiman2017globally}
Jonathan Raiman and John Miller. 2017.
\newblock Globally normalized reader.
\newblock In \emph{EMNLP}.

\bibitem[{Raison et~al.(2018)Raison, Mazar{\'e}, Das, and
  Bordes}]{raison2018weaver}
Martin Raison, Pierre-Emmanuel Mazar{\'e}, Rajarshi Das, and Antoine Bordes.
  2018.
\newblock Weaver: Deep co-encoding of questions and documents for machine
  reading.
\newblock \emph{arXiv preprint arXiv:1804.10490}.

\bibitem[{Rajpurkar et~al.(2016)Rajpurkar, Zhang, Lopyrev, and Liang}]{squad}
Pranav Rajpurkar, Jian Zhang, Konstantin Lopyrev, and Percy Liang. 2016.
\newblock Squad: 100,000+ questions for machine comprehension of text.
\newblock In \emph{EMNLP}.

\bibitem[{Seo et~al.(2017)Seo, Kembhavi, Farhadi, and Hajishirzi}]{bidaf}
Minjoon Seo, Aniruddha Kembhavi, Ali Farhadi, and Hannaneh Hajishirzi. 2017.
\newblock Bidirectional attention flow for machine comprehension.
\newblock In \emph{ICLR}.

\bibitem[{Seo et~al.(2018)Seo, Kwiatkowski, Parikh, Farhadi, and
  Hajishirzi}]{piqa}
Minjoon Seo, Tom Kwiatkowski, Ankur Parikh, Ali Farhadi, and Hannaneh
  Hajishirzi. 2018.
\newblock Phrase-indexed question answering: A new challenge for scalable
  document comprehension.
\newblock In \emph{EMNLP}.

\bibitem[{Shen et~al.(2017)Shen, Huang, Gao, and Chen}]{shen2017reasonet}
Yelong Shen, Po-Sen Huang, Jianfeng Gao, and Weizhu Chen. 2017.
\newblock Reasonet: Learning to stop reading in machine comprehension.
\newblock In \emph{KDD}.

\bibitem[{Shrivastava and Li(2014)}]{alsh2}
Anshumali Shrivastava and Ping Li. 2014.
\newblock Asymmetric lsh (alsh) for sublinear time maximum inner product search
  (mips).
\newblock In \emph{NIPS}.

\bibitem[{Voorhees and Tice(2000)}]{trecqa}
Ellen~M Voorhees and Dawn~M Tice. 2000.
\newblock Building a question answering test collection.
\newblock In \emph{SIGIR}.

\bibitem[{Wang et~al.(2018{\natexlab{a}})Wang, Yu, Guo, Wang, Klinger, Zhang,
  Chang, Tesauro, Zhou, and Jiang}]{wang2017r}
Shuohang Wang, Mo~Yu, Xiaoxiao Guo, Zhiguo Wang, Tim Klinger, Wei Zhang, Shiyu
  Chang, Gerry Tesauro, Bowen Zhou, and Jing Jiang. 2018{\natexlab{a}}.
\newblock R 3: Reinforced ranker-reader for open-domain question answering.
\newblock In \emph{AAAI}.

\bibitem[{Wang et~al.(2018{\natexlab{b}})Wang, Yu, Jiang, Zhang, Guo, Chang,
  Wang, Klinger, Tesauro, and Campbell}]{wang2017evidence}
Shuohang Wang, Mo~Yu, Jing Jiang, Wei Zhang, Xiaoxiao Guo, Shiyu Chang, Zhiguo
  Wang, Tim Klinger, Gerald Tesauro, and Murray Campbell. 2018{\natexlab{b}}.
\newblock Evidence aggregation for answer re-ranking in open-domain question
  answering.
\newblock In \emph{ICLR}.

\bibitem[{Weston et~al.(2015)Weston, Chopra, and Bordes}]{weston2014memory}
Jason Weston, Sumit Chopra, and Antoine Bordes. 2015.
\newblock Memory networks.
\newblock In \emph{ICLR}.

\bibitem[{Yang et~al.(2019)Yang, Xie, Lin, Li, Tan, Xiong, Li, and
  Lin}]{yang2019end}
Wei Yang, Yuqing Xie, Aileen Lin, Xingyu Li, Luchen Tan, Kun Xiong, Ming Li,
  and Jimmy Lin. 2019.
\newblock End-to-end open-domain question answering with bertserini.
\newblock \emph{arXiv preprint arXiv:1902.01718}.

\bibitem[{Yang et~al.(2015)Yang, Yih, and Meek}]{wikiqa}
Yi~Yang, Wen-tau Yih, and Christopher Meek. 2015.
\newblock Wikiqa: A challenge dataset for open-domain question answering.
\newblock In \emph{EMNLP}.

\end{thebibliography}
\bibliographystyle{acl_natbib}

\clearpage
\onecolumn
\appendix
% \section{Non-BERT Text Encoder}
% \input{90-nobert.tex}
\section{More Prediction Samples}\label{sec:sample}
\begin{table}[ht!]
    \centering
    \resizebox{\columnwidth}{!}{%
    \begin{tabular}{ll}
        \toprule
    
        % \midrule
        % \multicolumn{2}{l}{Q: What was Germany's central interest?}\\
        % \multicolumn{2}{l}{A: Europe} \\
        % \midrule
        % DrQA & [\textit{French Third Republic}] \textbf{Foreign policy} was of central interest to France during the ... \\
        % & [\textit{Imperialism}] ... over colonies would distract Germany from its central interest, \textbf{Europe itself}. \\
        % & [\textit{Otto von Bismarck}] ...distract Germany from its central interest, \textbf{Europe itself}. \\
        % \ours & [\textit{Geopolitik}] Germany was concerned primarily with \textbf{Eastern Europe}, and Italy's natural ...  \\
        % & [\textit{Military history of Finland during World War II}] The main strategic interest of Germany \\
        % & in the region were \textbf{the nickel mines in the Petsamo area}. \\
        % & [\textit{Blockade of Germany (1939–45)}] ... but Hitler's main problem was \textbf{oil}, around 12.5m tons ... \\
        
        \multicolumn{2}{l}{Q: Who became the King of the Canary Islands?}\\
        \multicolumn{2}{l}{A: Bethencourt} \\
        \midrule
        DrQA & [\textit{Canary Islands}] ... \textbf{Winston Churchill} prepared plans (...) of the Canary Islands ... \\
        & [\textit{Isleño} ... In 1501, \textbf{Nicolás de Ovando} left the Canary Islands ...\\
        & [\textit{Canary Islands}] ... over by \textbf{Fernando Clavijo}, the current President of the Canary Islands ... \\
        \ours & [\textit{Tenerife}] In 1464,\textbf{ Diego Garcia de Herrera}, Lord of the Canary Islands, ...  \\
        & [\textit{Bettencourt}] ... explorer \textbf{Jean de Béthencourt}, who conquered the Canary Islands ... \\
        & [\textit{Bettencourt}] ... \textbf{Jean de Béthencourt}, organized an expedition to conquer the Canary Islands, ... \\
        
        \midrule
        \multicolumn{2}{l}{Q: When was the outbreak of World War I?}\\
        \multicolumn{2}{l}{A: August 1914} \\
        \midrule
        DrQA & [\textit{Australian Army during World War II}] ... following the outbreak of war in \textbf{1939} and ... \\
        & [\textit{Australian Army during World War II}] ... The result was that when war came in \textbf{1939}, ... \\
        & [\textit{Australian Army during World War II}] ... the outbreak of the Korean War on 25 June \textbf{1950} ... \\
        \ours & [\textit{SMS Kaiser Friedrich III}] ... the outbreak of World War I in \textbf{July 1914}. \\
        & [\textit{Germany at the Summer Olympics}] At the outbreak of World War I in \textbf{1914}, organization ... \\
        & [\textit{Carl Hans Lody}] ... outbreak of the First World War on \textbf{28 July 1914} resulted in ... \\
        
        \midrule
        \multicolumn{2}{l}{Q: What comedian is also a university graduate?}\\
        \multicolumn{2}{l}{A: Mike Nichols} \\
        \midrule
        DrQA & [\textit{Anaheim University}] ... winning actress and comedian \textbf{Carol Burnett} in memory ...\\
        & [\textit{Kettering University}] \textbf{Bob Kagle} (...) is one of the most successful venture capitalists ... \\
        & [\textit{Kettering University}] \textbf{Edward Davies} (...) is the father-in-law of Mitt Romney.\\
        \ours & [\textit{University of Washington}] ... and actor and comedian \textbf{Joel McHale} (1995, MFA 2000). \\
        & [\textit{Michigan State University}] ... Fawcett; comedian \textbf{Dick Martin, comedian Jackie Martling} ... \\
        & [\textit{West Virginia State University}] ... a comedy show by famed comedian, \textbf{Dick Gregory}. \\
        
        \midrule
        \multicolumn{2}{l}{Q: Who is parodied on programs such as Saturday Night Live and The Simpsons?}\\
        \multicolumn{2}{l}{A: Doctor Who fandom} \\
        \midrule
        DrQA & [\textit{The Last Voyage of the Starship Enterprise}] ... the ``Saturday Night Live'' parody of \\
        & ``Star Trek'' with \textbf{William Shatner}, ... \\
        & [\textit{Saturday Night Live}] ... ``Saturday Night Live with \textbf{Howard Cosell}'' on the rival network ... \\
        & [\textit{Fox Broadcasting Company}] ... ``The Late Show'', which was hosted by comedian \textbf{Joan Rivers}. \\
        \ours & [\textit{Gilda Radner}] ... and ``Baba Wawa'', a parody of \textbf{Barbara Walters}. \\
        & [\textit{This American Life}] ...  Armisen parodied \textbf{Ira Glass} for a skit on ``Saturday Night Live''s ... \\
        & [\textit{Anton Chigurh}]... \textbf{Chigurh} has been parodied in other media, mainly as a spoof ... \\
        
        \bottomrule
    \end{tabular}
    }
    \caption{More prediction samples from DrQA and \ours. Each sample shows [\textit{document title}], context, and \textbf{predicted answer}.}
    \label{tab:od-more-samples}
\end{table}

\end{document}